\documentclass[letterpaper, 10 pt, conference]{ieeeconf}  %

\IEEEoverridecommandlockouts                              %

\usepackage{amsmath} %
\usepackage{amssymb}  %
\usepackage{graphicx}
\usepackage{subfigure}
\usepackage{units}
\usepackage{multirow}
\usepackage{array}
\usepackage{color}
\usepackage{url}
\usepackage{tabularx,ragged2e}

\newcolumntype{C}{>{\Centering\arraybackslash}X} %

\usepackage{mathtools}

\DeclarePairedDelimiter\floor{\lfloor}{\rfloor}

\title{\LARGE \bf
Learned Uncertainty Calibration for Visual Inertial Localization
}

\author{Stephanie Tsuei$^{1}$, Stefano Soatto$^{1}$, Paulo Tabuada$^{2}$, and Mark B. Milam$^{3}$%
\thanks{$^{1}$UCLA Department of Computer Science, Los Angeles, CA
        {\tt\small [stephanietsuei,soatto]@cs.ucla.edu}}%
\thanks{$^{2}$UCLA Department of Electrical Engineering, Los Angeles, CA
        {\tt\small tabuada@ee.ucla.edu}}%
\thanks{$^{3}$NG Next, Northrop Grumman, Redondo Beach, CA
        {\tt\small mark.milam@ngc.com}}%
}

\begin{document}

\maketitle
\thispagestyle{empty}
\pagestyle{empty}

\begin{abstract}
The widely-used Extended Kalman Filter (EKF) provides a straightforward recipe to estimate the mean and covariance of the state given all past measurements in a causal and recursive fashion. For a wide variety of applications, the EKF is known to produce accurate estimates of the mean and typically inaccurate estimates of the covariance. For applications in visual inertial localization, we show that inaccuracies in the covariance estimates are \emph{systematic}, i.e. it is possible to learn a nonlinear map from the empirical ground truth to the estimated one. This is demonstrated on both a standard EKF in simulation and a Visual Inertial Odometry system on real-world data.
\end{abstract}

\section{Introduction}

While deep learning has dominated much of the computer vision literature in recent years, ``traditional'' filtering methods still perform better in localization problems that use one or more cameras, such as Visual Odometry (VO) and Visual Inertial Odometry (VIO). This is because the filters hard-code known nonlinear kinematic models that are difficult to learn from data, although deep neural networks are often used to learn feature representations of measured data. The Extended Kalman Filter (EKF) is the most common  because it is so simple, even though complex non-linearities in the rotational dynamics and patent violation of the Gaussian assumption in the visual measurements remove all guarantees of convergence and accuracy. In practice, the EKF provides accurate state estimates $\hat x$ and overconfident covariance estimates $\hat P$ \cite{bailey2006consistency}. To improve the covariance estimates of VO problems, \cite{vega-brown_cello_2013} and \cite{liu_deep_2018} learn time-dependent measurement noise covariances $Q$ while \cite{de_maio_simultaneously_2020} uses a deep convolutional network to correct $\hat x$ and $\hat P$ directly from images. On the other hand, \cite{huang_first-estimates_2009, kottas_consistency_2013,heo_consistent_2018, hesch_camera-imu-based_2014, brossard_exploiting_2019, zhang_convergence_2017, forster_-manifold_2017} improve VIO covariance estimation by using filters specific to VIO. The accuracy of covariance estimates are evaluated by tabulating the percentage of timesteps where the estimation error is within the 1,2,3-$\sigma$ bounds dictated by the estimated covariance $\hat P$.

Our approach is most similar to \cite{de_maio_simultaneously_2020} in that we use supervised machine learning to improve the covariance estimates, but the input to our models only consist of $\hat P$ and $\hat x$ instead of the entire input image; having only $\hat P$ and $\hat x$ as inputs allows us to use much smaller networks. We also create a statistical test similar to that used in \cite{huang_first-estimates_2009,forster_-manifold_2017} to evaluate the calibration of estimates from a state estimator, except without the need for Monte-Carlo experiments (Sect.~\ref{sec:eval_calib}). Using that statistical test, we show that ground-truth covariance can be computed assuming ergodicity when Monte-Carlo experiments are impractical (Section \ref{sec:indcalib}) and also count the number of timesteps that are within 1,2,3-$\sigma$ bounds given by $\hat P$. In Section \ref{sec:illustrations}, we test our method on two simple examples as a quick validation. Section \ref{sec:vio} contains our main experiment, where we test our method on a VIO system processing real-world data. We achieved significant calibration using both a state-independent model $\phi(\hat P)$ and a state-dependent model $\phi(\hat x)$, which implies that the state $\hat x$ contains very little information about the miscalibration of $\hat P$ -- most of the miscalibration of $\hat P$ can be computed from $\hat P$ itself.

\section{Evaluating Calibration of Kalman Filters} \label{sec:eval_calib}

\subsection{Background: Sources of Error in EKFs}
Consider a nonlinear discrete-time dynamical system and measurement model with state $x \in \mathbb R^n$ and measurement $y \in \mathbb R^m$:
\begin{equation}
\begin{aligned}
    x_{k} &= f(x_{k-1},u_{k-1}) + \nu_k \\
    y_k &= h(x_k) + w_k
    \label{eq:dynamicalsys}
\end{aligned}
\end{equation}
Assume that $\nu_k \sim \mathcal N(0,R)$ and $w_k \sim \mathcal N(0,Q)$ are white Gaussian noise processes, and the input $u_k$ is known, along with the dynamics $f$ and nominal measurement model $h$. An Extended Kalman Filter (EKF) recursively computes an estimate of $x_k$, $\hat x_k$, along with its covariance  $\hat P_k = {\mathbb E}\left[(x_k-\hat x_k)(x_k-\hat x_k)^T \right]$  whenever it receives a new measurement $y_k$ by computing the quantities:
\begin{equation}
\begin{aligned}
    \hat x_{k|k-1} &= f(\hat x_{k-1},u_{k-1}) \\
    \hat P_{k|k-1} &= A_k \hat P_{k-1} A_k^\top + R \\
    K &= \hat P_{k|k-1} C_k^\top (C \hat P_{k|k-1} C^\top + Q)^{-1} \\
    \hat x_k &= \hat x_{k|k-1} + K (y_k - h(\hat x_{k|k-1})) \\
    \hat P_k &= (I - KC) \hat P_{k|k-1} (I-KC)^\top  + K Q K^\top.
    \label{eq:ekf}
\end{aligned}
\end{equation}
The matrices $A_k$ and $C_k$ are the Jacobians of $f(x,u)$ and $h(x)$ with respect to $x$ evaluated at $\hat x_{k-1}$ and $u_{k-1}$. Estimates $\hat x_k$ and $\hat P_k$ represent a posterior Gaussian distribution. If $f(x,u)$ and $h(x)$ are both linear, then as $k$ increases, $\hat x_k$ is guaranteed to converge to $x_k$ and the computation of the $\hat P_k$ are completely separate from the computation of the $\hat x_k$. Moreover, the innovation $z_k=y_k-h(\hat x_{k|k-1})$ should be zero-mean and white in both components and time. These same guarantees do not apply when either $f(x,u)$ or $h(x)$ are nonlinear. In the nonlinear case, the mean and innovation are computed using the original nonlinear model $f$ and the covariance is updated using linearized models. These unaccounted linearization errors mean that $\hat P_k$ is usually underestimated \cite{bailey2006consistency}. Finally, most implementations of VO and VIO treat $Q$ and $R$ as constants, though they may be state and time-dependent.

\subsection{1,2,3-$\sigma$ Intervals in Multiple Dimensions}
For a set of discrete samples $\phi_k$, $k=1,\dots,N$ drawn from a 1-D Gaussian distribution $\mathcal N(\mu_k,\sigma_k)$, about 68\% lie in the interval $\mu_k \pm \sigma_k$, 95\% in $\mu_k \pm 2\sigma_k$, and 99.7\%  in $\mu_k \pm 3\sigma_k$. %
The same can be done for a set of points $v_k \in \mathbb R^d$, each from a potentially different multivariate Gaussian distribution $\mathcal N(u_k, \Sigma_k)$. First, diagonalize each $\Sigma_k$ with eigenvalue decomposition: $\Sigma_k = X_k \Lambda_k X_k^\top$. Then, the columns of the matrix $X_k \Lambda_k^{1/2}$ form an orthogonal, but not orthonormal, basis and
\begin{equation}
    \nu_k = (X_k \Lambda_k^{1/2})^{-1} (v_k - u_k)
    \label{eq:newbasis}
\end{equation}
contains the coordinates of $v_k$ in the new coordinate system. Then, for each dimension $1, \dots, d$ of $\nu_k$, 68\% of samples are in the interval $[-1,1]$, 95\%  in  $[-2,2]$ and 99.7\% in $[-3,3]$. By counting the value of $\nu_k$ for each dimension at each timestep, we can evaluate the filter's calibration for each individual dimension, but not overall.

\subsection{Overall Calibration with Monte-Carlo Simulations} \label{sec:MCcalib}

Let $e_k = x_k - \hat x_k$ be the estimation error and $\hat \rho_k$ be the normalized estimation error squared (NEES), or square of the Mahalanobis distance at each timestep $k$:
\begin{equation}
    \hat \rho_k = e_k^\top \hat P_k^{-1} e_k
\label{eq:SolveSigma}
\end{equation}
$\hat \rho_k \sim \chi^2_n$, i.e. $\hat \rho_k$ is a $\chi^2$ variable with $n$ degrees of freedom. If we run $M$ Monte-Carlo simulations and compute a value of $\rho_{k,i}$ for every timestep $k$ in each run $i$, then their sum $\hat{\bar \rho}_k = \sum_{i=1}^M \hat \rho_{k,i}$ is a $\chi^2$ variable with $M \times n$ degrees of freedom. Then, over the Monte-Carlo runs, we can compute confidence intervals for values of $\hat{\bar \rho}_k$. Values for $\hat{\bar \rho}_k$ should remain within the confidence interval for all $k$ if the $\hat P_k$ are well-calibrated. If $\hat{\bar \rho}_k$ is consistently too high, then the covariance estimates $\hat P_k$ are too optimistic. If $\hat{\bar \rho}_k$ is consistently too low, then the $\hat P_k$ are conservative. This approach was used in \cite{huang_first-estimates_2009, forster_-manifold_2017} to measure the accuracy of their covariance estimates.

Unfortunately, Monte-Carlo approximations are not scalable, and not practical in a real-world scenario where one would have to carefully place the sensor platform in the same precise position and orientation at every run to conduct repeated trials. %
Also, to perform repeated and sufficiently exciting motions required for VIO, one would need a precise actuation system, like a robot arm and not a quadcopter. Therefore, we need another approach to evaluate the covariance calibration.

\subsection{Exploiting Residual Independence for Calibration} \label{sec:indcalib}

Here, we present a finer-grained method for evaluating calibration in a multiple dimensions with a goodness-of-fit test for unbiased estimators that does not require Monte-Carlo simulations. First, we assume that the $e_k$ are approximately independent. While not strictly satisfied, this assumption enables a practical procedure, which we will then validate empirically. Next, let $p_{\hat \rho_k}$ be the approximate probability density function of $\hat \rho_k$, which can be computed with a normalized histogram of the $\hat \rho_k$. Next, since each $\hat \rho_k \sim \chi^2_n$ if the system is well-calibrated, we can then use $p_{\hat \rho_k}(x)$ in a goodness-of-fit test with the $\chi^2_n$ distribution. For this work, we use the L2 divergence \cite{poczos_nonparametric_2011} between this approximate density and the density of $\chi^2_n$ as a comparison metric, $p_{\chi^2_n}(x)$:
\begin{equation}
    \mathcal D_{L_2}(\hat \rho_k \| \chi^2_n) = \left ( \int_0^\infty (p_{\hat \rho_k}(x) - p_{\chi^2_n}(x))^2 dx \right )^{1/2}
\label{eq:divergence}
\end{equation}
$\mathcal D_{L_2}$ is easy to compute and useful for comparing goodness-of-fit of multiple sets of $\hat \rho_k$ on the same dataset, but not as an absolute measure as one would use a p-value. We will use it to compare methods of calibration to ground truth covariance in the rest of the paper.

\section{Computing a Calibrated Covariance} \label{sec:hypotheses}
Our hypothesis is that the covariance estimates provided by an EKF present systematic errors and because of that, there exists a learnable map from the estimated value to a more calibrated value that can be executed in real-time. We consider maps of the following forms, in order of complexity:
\begin{enumerate}
    \item A multiplicative scalar: All components of the estimated covariance are offset by a single scaling factor, $P_k = s \hat P_k$. %
    \item $P_k = A \hat P_k A^\top$, i.e. the map is a constant  transformation of the covariance. %
    \item $Q_k = \phi(\hat P_k)$ and $P_k = Q_k Q_k^\top$, i.e. the map is an arbitrary function of the estimated covariance. This map, and the next, can be implemented by a feedforward neural network. %
    \item $Q_k = \phi(\hat x_k, \hat P_k)$ and $P_k = Q_k Q_k^\top$, i.e. the map is an arbitrary function of the estimated state and the estimated covariance. %
\end{enumerate}
An even more general model would be a map represented with a recurrent neural network. Our experiments show, however, that the memoryless maps of hypotheses 3 and 4 are sufficient for uncertainty calibration.

\subsection{Finding Ground-Truth Covariance}

Validating any of the hypotheses above requires a ``ground-truth" value of covariance. In a Monte-Carlo experiment, we can use the unbiased sample covariance at a given timestep $k$ given measurements $e_{k,i}$:
\begin{equation}
    \tilde P_k = \frac{1}{M-1} \sum_{i=1}^M e_{k,i} e_{k,i}^\top
    \label{eq:GTMonteCarlo}
\end{equation}

However, running many nearly identical tests on real-world equipment to measure ground-truth covariance is costly and time-consuming. For real-world experiments, we can compute a pseudo-ground-truth covariance with only one test if we additionally assume that that the motion state is approximately \emph{ergodic}, i.e. that the population statistics match the temporal statistics. Practically, assuming ergodicity means assuming that errors in the motion state vary slowly over time, which is often true for converged filters. For an odd-sized time window $K$, we define pseudo-ground-truth as
\begin{equation}
    \tilde P_k = \frac{1}{K-1} \sum_{k-\floor{K/2}}^{k+\floor{K/2}} e_{k} e_{k}^\top.
    \label{eq:GTSample}
\end{equation}
We can then use $\tilde P_k$ instead of $\hat P_k$ in \eqref{eq:SolveSigma} to compute a new set of $\tilde \rho_k$ and a new value for $\mathcal D_{L_2}$. For simplicity, we discard timesteps for which we cannot compute a sample covariance, $k \le \floor{K/2}$ and $k \ge N-\floor{K/2}$, where $N$ is the total number of timesteps, from further analysis. Using these $\tilde \rho_k$, we can verify the ergodic assumption: if the ergodic assumption is true, then $\mathcal D_{L_2}(\tilde \rho_k \| \chi^2_n)$ should be small. In the experiments section, we visualize typical ``small" and ``large" values of $\mathcal D_{L_2}$.

The authors of \cite{liu_deep_2018} and \cite{de_maio_simultaneously_2020} trained neural networks to predict covariances by training them with negative log-likelihood (NLL) losses. Unlike the typical maximum likelihood problem in which a few parameters are estimated from many data, in their NLL loss every datum is potentially drawn from a different distribution. We believe that they were nevertheless able to train a neural network to predict covariances because their training data was approximately ergodic and therefore not drawn from many different distributions.

\subsection{Hypothesis 1: Constant Multiplicative Scalar}

We solve for a constant factor over all $M$ sequences in a training dataset using the following optimization problem: 
\begin{equation}
\begin{aligned}
& \underset{s}{\text{minimize}}
& & \sum_{q=1}^{M} \sum_{k=0}^{N} \sum_{i=0}^n \sum_{j=i}^n (s \hat P_{q,k|k}^{i,j} - P_{q,k}^{i,j})^2 \\
& \text{subject to}
& & s \geq 0 
\end{aligned}
\label{eq:globalscalaropt}
\end{equation}
where $P^{i,j}_{q,k}$ denotes the $i,j$th entry of the covariance matrix from the $k$th timestep of the $q$th sequence. This hypothesis is  simplistic and not powerful enough, but it is an easily solvable quadratic program.

\subsection{Hypothesis 2: Constant Linear Transformation}
The second hypothesis is expressed as the optimization problem with decision variable $A \in \mathbb R^{n \times n}$

\begin{equation}
\begin{aligned}
& \underset{A}{\text{minimize}}
& & \sum_{q=1}^{M} \sum_{k=0}^{N} \sum_{i=0}^n \sum_{j=i}^n ((A \hat P_{q,k|k} A^\top - P_{q,k})^{i,j})^2 \\
\end{aligned}
\label{eq:globallinearopt}
\end{equation}
This quartic and nonconvex optimization problem is the natural next step from Hypothesis 1. We implement Hypothesis 2 using IPOPT \cite{wachter_implementation_2006}. Theoretically, this adjustment should be at least as effective as the constant multiplicative scalar, since $A=sI$, where $s$ is the solution to \eqref{eq:globalscalaropt}, is a feasible solution to \eqref{eq:globallinearopt}. However, because local nonconvex optimizers are sensitive to the initial guess and are not guaranteed to return the correct solution, the calibration of Hypothesis 2 is not uniformly better than that of Hypothesis 1. In Sections \ref{sec:illustrations} and \ref{sec:vio}, it is always worse than Hypothesis 1.

\subsection{Hypothesis 3 and 4: Fully-Connected Neural Networks}

Let $Q \in \mathbb R^{n \times n}$ be the output of the neural network and let the adjusted covariance be $QQ^\top$. We use a weighted elementwise difference between the upper triangles of $Q Q^\top$ and $P$ as the loss function:
\begin{equation}
    \mathcal L = \sum_{i=0}^n \sum_{j=i}^n w_{ij} ((Q Q^\top)^{i,j} - P^{i,j})^2
\end{equation}
Each timestep of each sequence serves as a training point. The inputs to the network are the upper triangle of a covariance matrix. In each of our experiments, we trained many simple feedforward neural networks using the Adam Optimizer in Tensorflow with varying architectures, L2 regularization weights, and epochs; very little thought was given to the architectures  we tested. The outputs of these neural networks are well-calibrated and the best performing architecture for each experiment is detailed in Sections \ref{sec:illustrations} and \ref{sec:vio}.

\section{Two Contrasting Illustrations} \label{sec:illustrations}

Before presenting our main experiment in Section \ref{sec:vio}, we illustrate concepts from the previous two sections with a couple of easy-to-visualize examples.

\subsection{Illustration 1: Linear Kalman Filter} \label{sec:linearkf}
\begin{figure}
    \centering
    \subfigure[State Errors]{\includegraphics[width=1.6in]{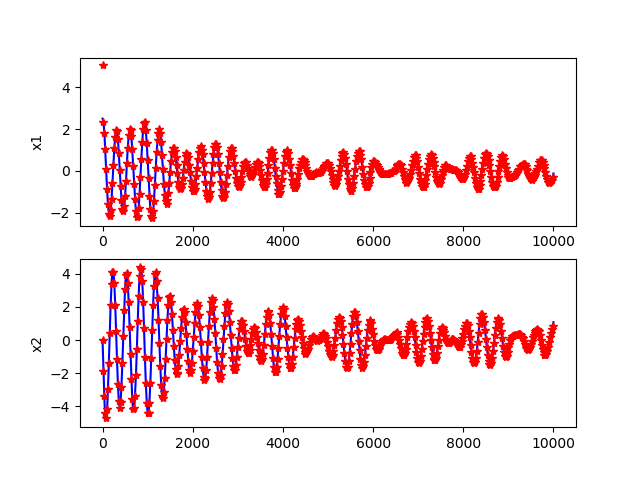}}
    \subfigure[Innovation]{\includegraphics[width=1.6in]{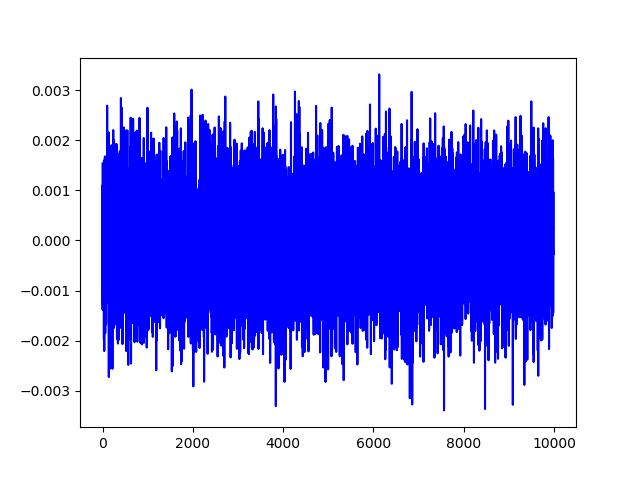}}
    \subfigure[$\chi^2_2$ Overlay with $p_{\hat \rho_k}$]{\includegraphics[width=1.6in]{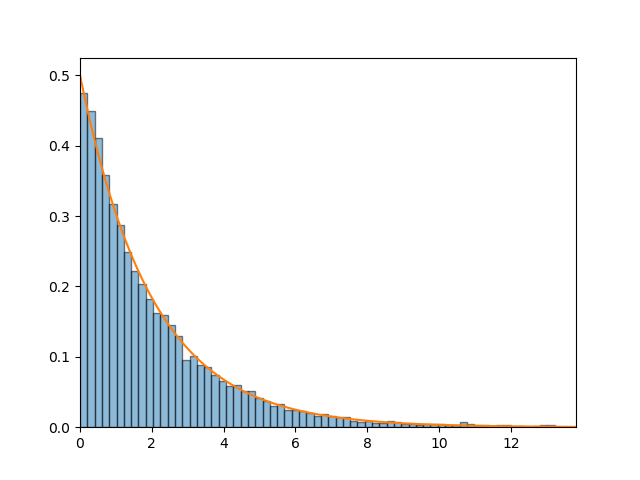}}
    \subfigure[$\chi^2_2$ Overlay with $p_{\rho_k}$]{\includegraphics[width=1.6in]{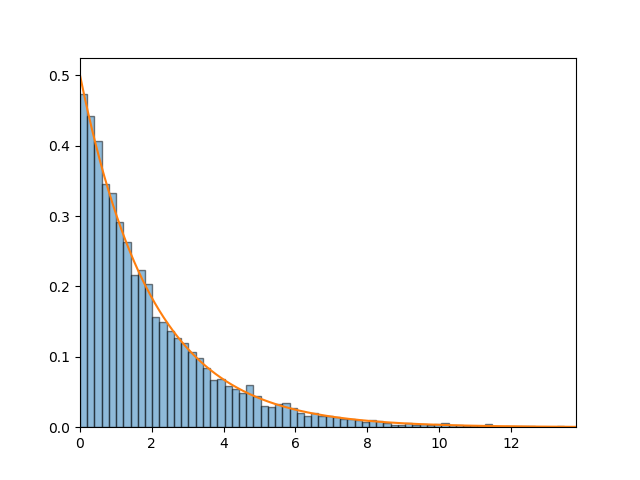}}
    \caption{The state estimation and innovation of the linear Kalman Filter for a single run are shown in (a) and (b) - the state estimation is accurate and the innovation is essentially white noise. (c) plots $p_{\bar \rho_k}$ against the $\chi^2_2$ density for a single run. Visually, the histogram and the $\chi^2_2$ density are very close, showing that the independence assumption holds and that the covariance estimates are well-calibrated. This is further verified in (d), which is the same plot as (c), except that the normalized histogram is computed using a ground-truth covariance from Monte-Carlo trials. \eqref{eq:GTMonteCarlo}.}
    \label{fig:linearkf_results}
\vspace{-1.2em}
\end{figure}

A system that should have perfectly calibrated state estimates is a spring-mass damper system with mass $m=1$, spring constant $k=4$, and damping coefficient $c=0.1$. After discretizing time into timesteps of length $\delta t = 0.01$s, the discrete-time dynamics and measurement equations are:
\begin{equation}
\begin{aligned}
    \begin{bmatrix} x_1 \\ x_2 \end{bmatrix}_{k+1} &= \begin{bmatrix} 1 & \delta t \\ -\frac{k}{m} \delta t & 1-\frac{c}{m}\delta t \end{bmatrix} \begin{bmatrix} x_1 \\ x_2 \end{bmatrix}_{k} + \begin{bmatrix} 0 \\ \delta t \end{bmatrix} u_k  + \nu_k \\
    y_k &= \begin{bmatrix} 1 & 0 \end{bmatrix}  \begin{bmatrix} x_1 \\ x_2 \end{bmatrix}_{k} + w_k
\end{aligned}
\end{equation}
where $x_1$ is the horizontal position, $x_2$ is the horizontal velocity, and $u$ is a forcing input. In our experiment, $u(t)$ is the discretized version of $\sin(\frac{\pi}{2}t)$. In both the simulation and the Kalman Filter, $\nu_k \sim \mathcal N (0, 0.003^2)$ and each element of $w_k \sim \mathcal N (0, 0.005^2)$. Results for a single run from a set of 50 Monte-Carlo trials are shown in Figure \ref{fig:linearkf_results}. The overlay of the $\chi^2_2$ density and the normalized histogram of the $\hat \rho_k$ are a near-perfect fit. Additionally, using the transformation in \eqref{eq:newbasis}, we find that in the first dimension 68.20, 95.42, and 99.74\% of points are within $\pm 1, 2, 3$, respectively. In the second dimension, the percentages are 68.12, 95.45, and 99.73\%.

\subsection{Illustration 2: EKF for 2D Localization} \label{sec:ekf}

\begin{figure}
    \centering
    \subfigure[State Errors]{\includegraphics[width=1.6in]{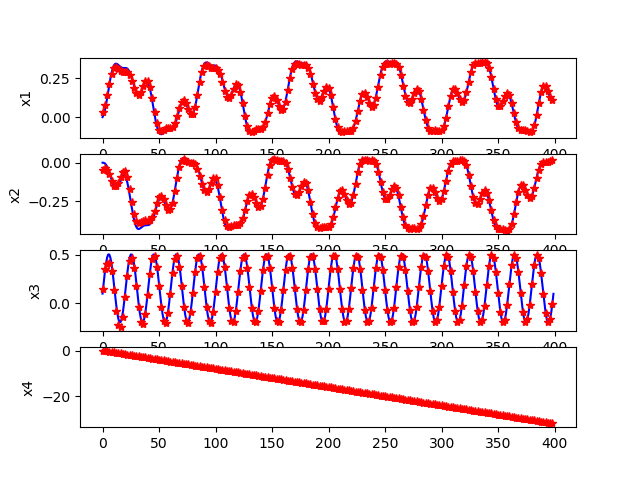}}
    \subfigure[Innovation]{\includegraphics[width=1.6in]{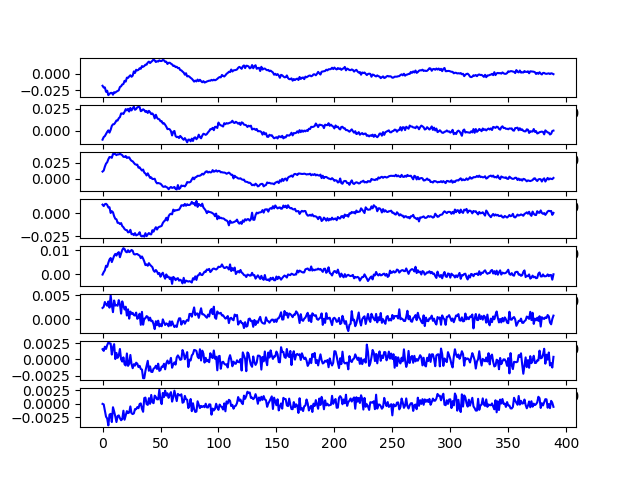}}
    \subfigure[$\chi^2_4$ Overlay with $p_{\hat \rho_k}$]{\includegraphics[width=1.6in]{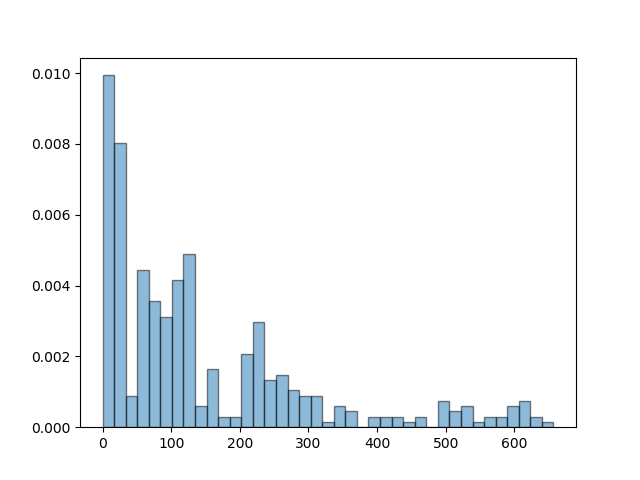}}
    \subfigure[$\chi^2_4$ Overlay with $p_{\rho_k}$]{\includegraphics[width=1.6in]{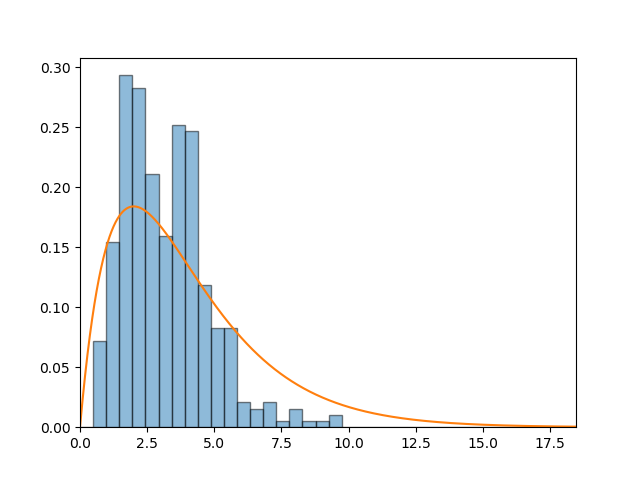}}
    \caption{Calibration results for the EKF's test sequence. The EKF has small estimation error (a) and poor covariance calibration. The innovation for this 2D localization problem (b) is clearly not white. In (c), the approximate density of $\rho_k$ is far from the $\chi^2_4$ density that we did not plot the $\chi^2_4$ pdf. Finally, in (d), the overlay is much closer to the ground-truth covariance computed using Monte-Carlo simulations, although still not a perfect fit because independence of the $e_k$ is only an approximation.}
    \label{fig:ekf_estcalib}
\end{figure}

We repeat the same process above for an extended Kalman Filter for a 2D localization problem. The system is a Dubin's car with state $[ x, y, v, \theta ]$ and acceleration $a$ and angular velocity $\omega$ inputs. It receives range and bearing measurements from a set of four known beacons located at $(3.5,-1.1), (10,10), (-5,15)$, and $(-10,-8.2)$. The discrete-time dynamics with discretization $\delta t = 0.1$s are:
\begin{equation}
\begin{aligned}
x_k &= x_{k-1} + \int_{t_{k-1}}^{t_k} v(\tau) \cos(\theta(\tau)) d\tau \\
y_k &= y_{k-1} + \int_{t_{k-1}}^{t_k} v(\tau) \sin(\theta(\tau)) d\tau \\
v_k &= a \delta t \\
\theta_k &= \omega \delta t
\end{aligned}
\end{equation}
and the measurement equations for beacon $i$ located at $[ x_i, y_i]$ are:
\begin{equation}
\begin{aligned}
r_{i,k} &= ( (x_i-x_k)^2 - (y_i-y_k)^2 )^{1/2} \\
\phi_{i,k} &= \arctan(y_i-y_k, x_i-x_k) - \theta
\end{aligned}
\end{equation}
In these experiments, we generate 11 training sequences and one test sequence of $a_k$ and $\omega_k$. In all 12 sequences $a_k$ is a sinusoid with frequency 0.5Hz and $\omega_k$ is a constant. Figure \ref{fig:ekf_estcalib} is the corresponding figure to Figure \ref{fig:linearkf_results} for this localization problem. It is clear that although the state estimation errors are small, the covariance estimates are inaccurate, and we test Hypotheses 1-4 on this problem. The neural network in Hypothesis 3 is a fully connected network with hidden layers that have 512, 512, 256, 256, 128, and 64 nodes. The L2 regularization weight was 1e-4 and it was trained for 50 epochs. The network for Hypothesis 4 is fully connected with five hidden layers that all have 128 nodes. Its L2 regularization weight was 1e-3 and was trained for 150 epochs. Both networks use ReLU activations, $w_{ij}=5$ on the diagonals, and $w_{ij}=1$ on the off-diagonals.

\begin{figure}
    \centering
    \subfigure[Scalar Adj. Overlay]{\includegraphics[width=1.6in]{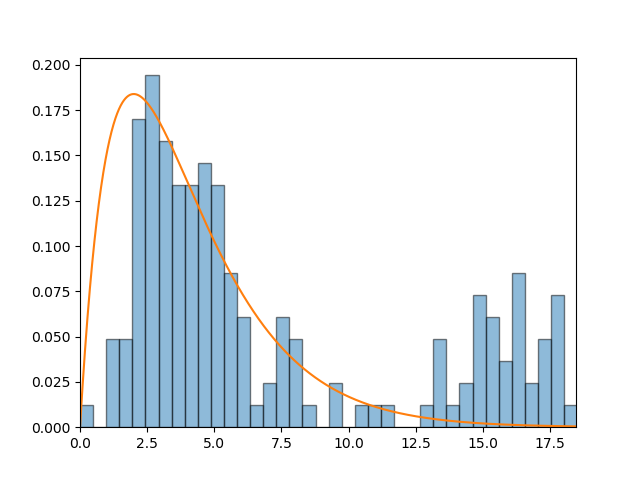}}
    \subfigure[Matrix Adj. Overlay]{\includegraphics[width=1.6in]{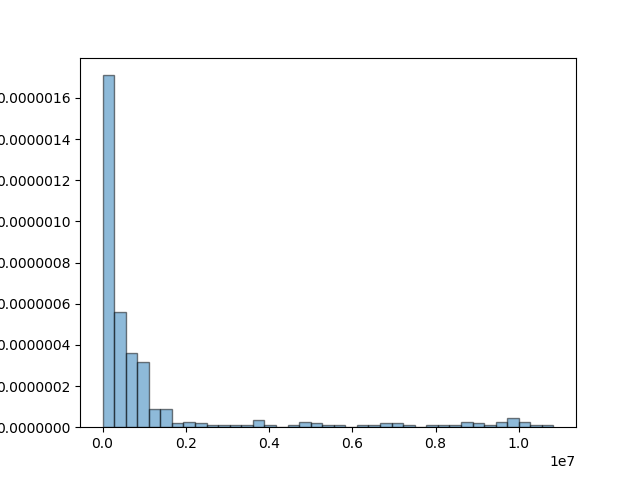}}
    \subfigure[NN Adj. Overlay]{\includegraphics[width=1.6in]{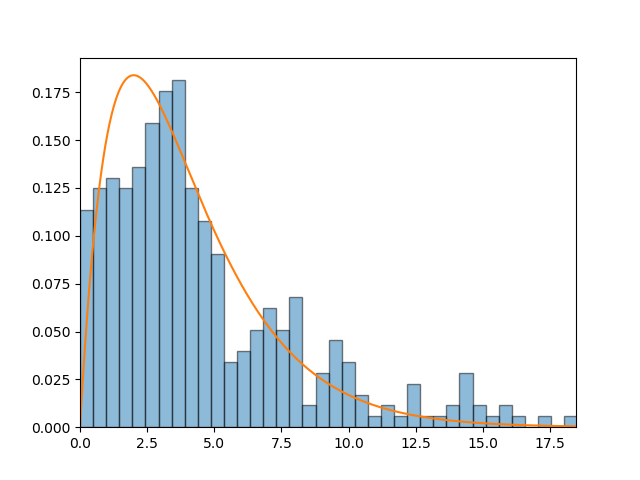}}
    \subfigure[NN w/State Overlay]{\includegraphics[width=1.6in]{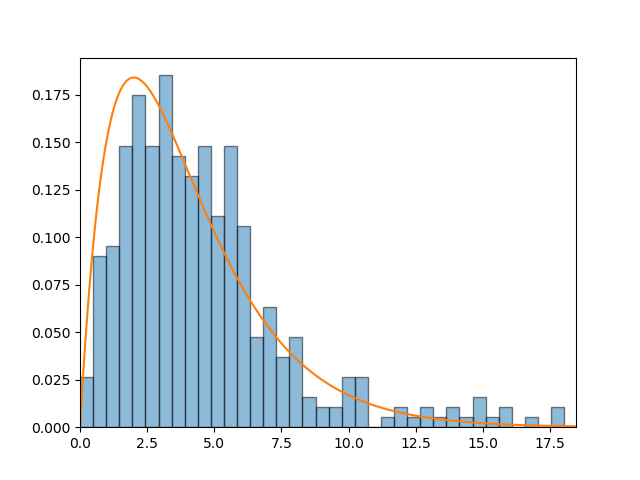}}
    \caption{Overlays of $\chi^2_4$ with $p_{\hat \rho_k}$ computed with adjusted covariances for the 2D localization problem. These overlays visualize the trends seen in Table \ref{tab:ekf_results}.}
    \label{fig:ekf_adjustments}
    \vspace{-0.4cm}
\end{figure}

Divergences for the unadjusted covariances, ground-truth, and the four hypotheses are shown in Table \ref{tab:ekf_results} with corresponding overlays are in Figures \ref{fig:ekf_estcalib} and \ref{fig:ekf_adjustments}. In Table \ref{tab:ekf_results}, the third column contains the mean and standard deviation of the divergences of 50 Monte-Carlo runs. Ground-truth covariances are computed using \eqref{eq:GTMonteCarlo} and the divergences in Table \ref{tab:ekf_results} are the mean and standard deviation of divergences of the 50 runs. The second column contains the reduction in divergence of the means in the second column as a percentage of the reduction from the unadjusted covariances to the ground-truth covariances. We observe that the calibration of the ground-truth covariances are within one standard deviations of the calibrations of both Hypothesis 3 and 4. The main conclusion of the results is that both Hypotheses 3 and 4 can achieve calibration, but that the state-dependence in Hypothesis 4 only yields a modest benefit. Note that in Figures \ref{fig:ekf_estcalib} and \ref{fig:ekf_adjustments}, we did not plot the $\chi^2$ density over the histogram when there was very little overlap between the two.

\begin{table*}
    \centering
    \caption{Calculated divergences for the 2D localization problem. }
    \begin{tabular}{|c|c|c|c|c|c|}
    \hline
                        & \textbf{\% Dec.} & \textbf{Test Set Sampling}  & \textbf{\% 1-}$\sigma$ & \textbf{\% 2-}$\sigma$ & \textbf{\% 3-}$\sigma$ \\
    \hline
    Original Estimated  & 0\%      & 0.3394 $\pm$ 0.0145   & 16.2, 17.6,  8.9, 58.5  &  30.0, 34.1, 17.1, 86.8  &  42.5, 48.3, 27.3, 94.8 \\
    \hline
    MC Ground-Truth     & 100\%    & 0.1839 $\pm$ 0.0547   & 60.8, 67.5, 68.5, 68.5  &  98.7, 96.1, 95.4, 95.9  &  99.7, 99.9, 99.5, 99.7 \\
    \hline
    Global Scalar       & 31.6\%   & 0.2902 $\pm$ 0.0510    & 30.2, 34.3, 17.3, 87.0  &  53.3, 60.3, 39.2, 96.7  &  69.3, 75.9, 60.8, 97.5 \\
    \hline
    Global Matrix       & -9.1\%   & 0.3535 $\pm$ 2.4e-06   & 5.6, 2.1, 0.7, 0.4      & 11.2, 4.1, 1.0, 0.5     & 16.2, 6.1, 1.4, 0.5 \\
    \hline
    Neural Network      & 75.4\%   & 0.2222 $\pm$ 0.0884    & 89.6, 62.0, 64.1, 45.6  & 98.0, 88.3, 88.0, 75.2  & 99.8, 96.3, 94.4, 89.0 \\
    \hline 
    State-Dependent NN  & 90.9\%   & 0.1981 $\pm$ 0.0553    & 95.7, 56.4, 51.1, 64.8  & 100.0, 95.3, 80.7, 91.4  & 100.0, 99.8, 92.7, 98.0 \\
    \hline
    \end{tabular}
    \label{tab:ekf_results}
\end{table*}

\section{Calibration of Visual Inertial Odometry} \label{sec:vio}

\begin{figure}
    \centering
    \subfigure[3D Trajectory]{\includegraphics[width=1.6in]{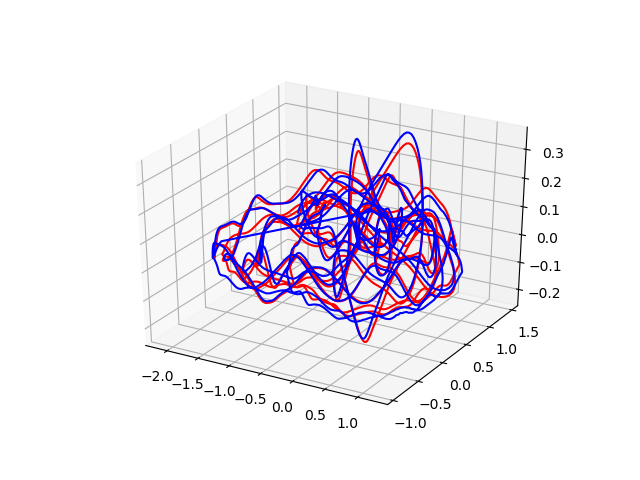}}
    \subfigure[Innovation]{\includegraphics[width=1.6in]{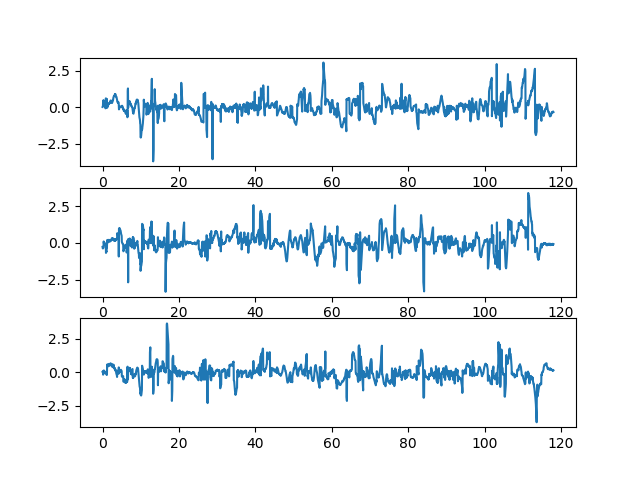}}
    \subfigure[$\chi^2_9$ Overlay]{\includegraphics[width=1.6in]{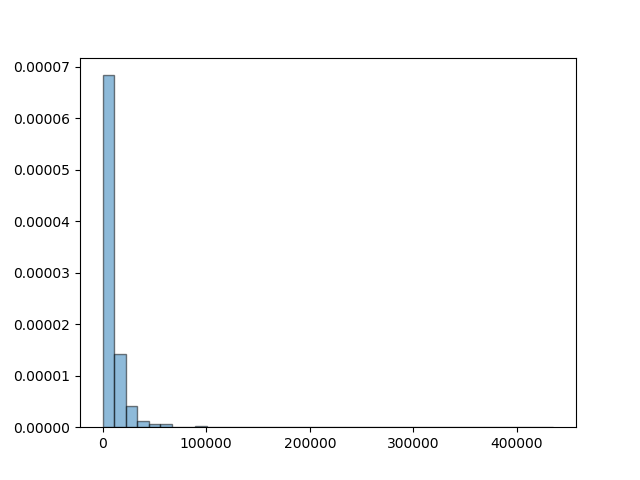}}
    \subfigure[Ground-Truth Overlay]{\includegraphics[width=1.6in]{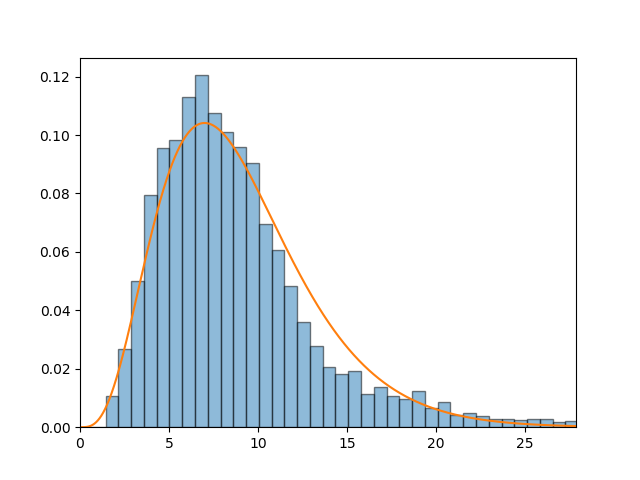}}
    \caption{The 3D trajectory, innovation, and overlays for the VIO test sequence. As with the EKF experiments, the state estimation error is small, but the innovation is clearly not white noise. In (c), there is very little overlap between the histogram approximation of $\hat \rho_k$ the $\chi^2_9$ distribution. (d) contains the same plot, except with $\hat \rho_k$ generated using ground-truth covariances computed using ergodicity from the test set. A visual comparison of (c) and (d) shows that the ergodic assumption and the independence assumption are both approximately true.}
    \label{fig:room6_traj}
    \vspace{-0.5cm}
\end{figure}

\begin{table*}[ht]
    \centering
    \caption{Table of computed divergences for the VIO experiment.}
    \begin{tabularx}{\textwidth}{|c|c|c|C|C|C|}
        \hline
                                & \textbf{\% Dec.} & \textbf{Test Set Sampling} & \textbf{\% 1-}$\sigma$ & \textbf{\% 2-}$\sigma$ & \textbf{\% 3-}$\sigma$\\
        \hline
        No Adjustment           & 0\%     & 0.2697 $\pm$ 5.61e-08 & 11.4,  9.5, 10.2,  6.2,  5.1,  6.0,  3.3,  2.4,  2.1 & 21.8, 19.1, 20.5, 12.0,  9.6, 12.9,  6.2,  4.7,  4.7 & 32.6, 29.1, 30.4, 18.2, 14.5, 19.2,  9.2,  7.3,  7.0 \\
        \hline
        ``Ground-Truth"         & 100\%   & 0.1103 $\pm$ 0.0223   & 60.4, 65.3, 67.4, 70.9, 71.7, 70.5, 67.1, 67.3, 70.3 & 98.2, 95.1, 96.1, 96.7, 96.9, 95.6, 95.8, 95.4, 95.1 & 99.9, 99.5, 99.5, 99.6, 99.7, 99.8, 99.7, 99.6, 99.4 \\
        \hline
        Global Scalar           & 42.7\%  & 0.1937 $\pm$ 0.0088   & 96.7, 93.4, 90.8, 74.7, 62.5, 79.7, 42.9, 30.1, 30.3 & 100.0, 97.8, 97.6, 95.5, 90.6, 96.8, 72.8, 50.3, 50.8 & 100.0, 98.6, 99.0, 98.8, 96.9, 99.0, 86.3, 65.1, 64.4 \\
        \hline
        Global Matrix           & 16.1\%  & 0.2433 $\pm$ 0.0037 & 98.1, 92.9, 78.1, 60.9, 66.5, 71.1, 60.0, 56.0, 74.2 & 99.7, 98.1, 83.0, 69.9, 77.0, 85.2, 71.1, 70.6, 90.3 & 99.9, 98.3, 83.9, 74.9, 82.8, 88.2, 75.3, 75.9, 94.9 \\
        \hline
        Neural Network          & 97.8\%  & 0.0987 $\pm$ 0.0159   & 76.4, 64.7, 72.8, 70.6, 70.6, 70.9, 69.8, 67.8, 65.7 & 96.5, 91.0, 93.3, 93.8, 94.0, 94.8, 93.2, 90.6, 90.1 & 99.6, 99.6, 98.5, 98.5, 98.8, 99.3, 98.4, 98.3, 98.1  \\
        \hline
        NN with State           & 105.6\% & 0.1028 $\pm$ 0.0219   & 69.9, 65.4, 70.5, 70.2, 71.7, 71.3, 71.1, 70.6, 71.3 & 95.8, 91.2, 94.9, 95.0, 95.9, 95.6, 94.2, 94.2, 94.8 &  99.1, 99.2, 99.6, 99.3, 99.2, 99.4, 99.0, 99.2, 99.3 \\
        \hline
    \end{tabularx}

    \label{tab:vio_adjustment_results}
\end{table*}

In the VIO problem, the state $x$ consists of the orientation, position, velocity, the map states, and any autocalibration states. Since we often do not have ground truth for the map, alignment, and autocalibration states, we will only analyze the localization states. Orientation is represented as a rotation vector, so in this work, the state dimension is $n=9$. A detailed description and derivation of the equations of motion for a VIO system can be found in \cite{jones_visual-inertial_2011}.

We evaluated XIVO, a reimplementation of the EKF SLAM system described in \cite{jones_visual-inertial_2011}, on the TUM Visual Inertial Dataset \cite{schubert_tum_2018}, a benchmark that features sequences of large, fast, and aggressive motions of a rig containing a stereo camera pair and an IMU. The dataset includes six sequences, named room1-room6, with ``ground-truth" position and orientation collected using a motion capture system. Since XIVO's algorithm only uses monocular images, and not stereo images, we effectively have twelve sequences, totalling 32,470 timesteps. XIVO's estimate of the position and orientation is comparable with other state-of-the-art VIO systems.\footnote{See the table at \url{https://github.com/ucla-vision/xivo/blob/devel/wiki.md} for XIVO's absolute trajectory error (ATE) and relative pose error (RPE). Note that these errors are not the same as the \emph{mean error} in Section \ref{sec:zeromean}.}

Since the motion capture system did not measure velocity, we backdifferenced the position ground truth in order to compute a velocity ground truth as well. Next, we used the method of Horn \cite{horn_closed-form_1987} to compute the transformation (a rotation and a translation) between the ground-truth points and the estimated points, since the two sets of points were not recorded in the same coordinate frame. The two room6 sequences were set aside as a test set while the other ten sequences were used for training. The trajectory of the test set, the innovation of the translation states, and overlay with the $\chi^2_9$ distribution are shown in Figure \ref{fig:room6_traj}.

\subsection{Validating the Zero-Mean Assumption} \label{sec:zeromean}

The analysis in Section \ref{sec:MCcalib} assumes that the errors are zero-mean. We compute the errors for all timesteps of the twelve sequences in the TUM VI dataset and find the mean of all of them. After interpolation and alignment of the ground-truth data, the mean translation, rotation, and velocity errors are 5.18e-17m, 0.0064rad, and 0.0017m/s, respectively while the mean Euclidean norms of ground-truth translation, rotation, and velocity across all 12 sequences in the dataset are 1.15m, 1.50rad, and 0.902m/s. The translation error is zero because the method of Horn optimizes translation error when computing the alignment between the coordinate frames from the ground-truth measurements and the estimated states. Although technically nonzero, the rotation and velocity errors are small when compared to the motions in the dataset. Therefore, we will consider them negligible and consider the mean of the errors to be zero.

\subsection{Validating the Ergodicity Assumption}

The sample covariance in \eqref{eq:GTSample} is computed for each timestep $k$ using a window of states centered around $k$. In order to find the best possible window size for each state of interest, we computed $\mathcal D_{L2}(\tilde \rho_k \| \chi^2)$ using odd-numbered window sizes between 27 and 601 for the ten training sequences. A window size of 275 produced low divergences for both the ten training sequences and the two test sequences. The divergence on the test set was 0.1105. The overlay with the $\chi^2_9$ distribution is in Figure \ref{fig:room6_traj}. Because these numbers are relatively small when compared to the divergences computed with the sampled covariances, and because of the relative visual fit compared to the overlay generated with the unadjusted covariance, we consider the ergodicity assumption validated.

\subsection{Experimental Results}

We run our ten sequences of training data through Hypotheses 1-4. For Hypotheses 3 and 4, we use ReLU activations, a L2 regularization weight of 0.001 in the loss function, and set $w_{ij}=10$ along the diagonals, $2.5$ for off-diagonals in the same state's 3 x 3 block, and 0.5 otherwise. The neural network for Hypothesis 3 had hidden layers with widths 1024, 512, 256, 128, 64 and was trained for 25 epochs. The best network for Hypothesis 4 had hidden layers with widths 256, 256, 256, 128, 128 and was trained for 50 epochs.

\begin{figure}
    \centering
    \subfigure[Scalar Adjustment]{\includegraphics[width=1.6in]{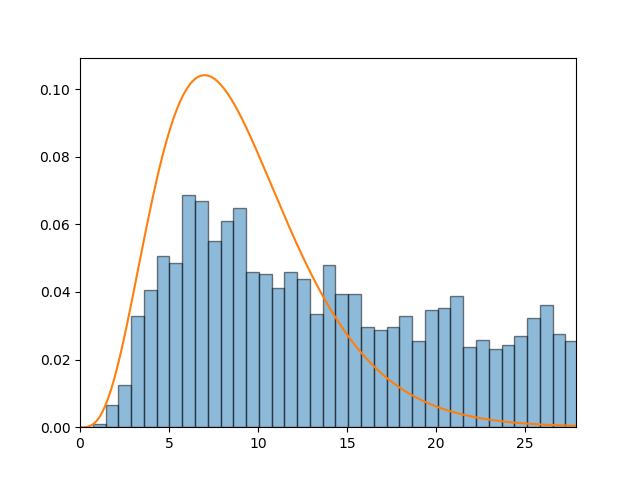}}
    \subfigure[Matrix Adjustmnet]{\includegraphics[width=1.6in]{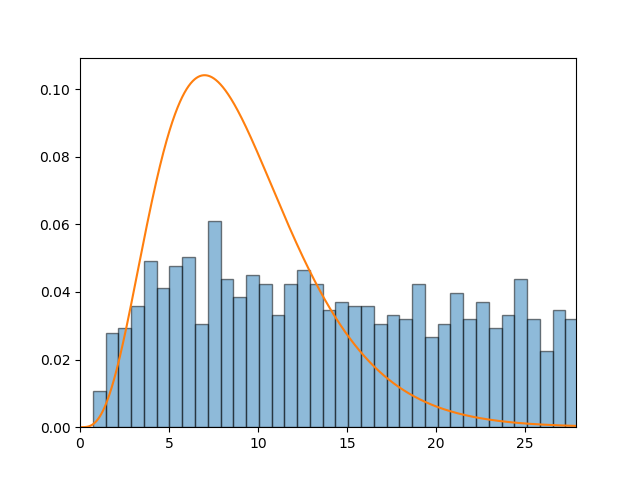}}
    \subfigure[Neural Network Adjustment]{\includegraphics[width=1.6in]{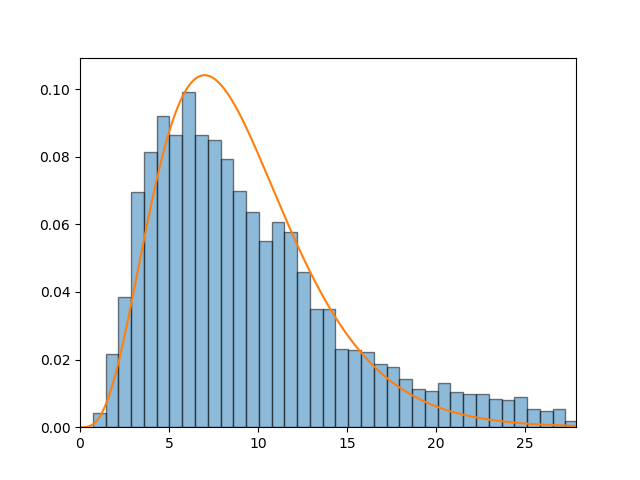}}
    \subfigure[State-Dependent NN Adjustment]{\includegraphics[width=1.6in]{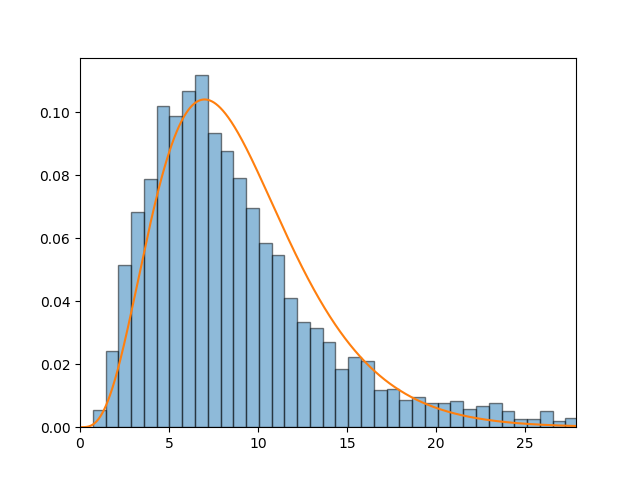}}
    \caption{Overlays of the test set's $\hat \rho_k$ computed with adjusted covariances - a visualization of the results in Table \ref{tab:vio_adjustment_results}.}

    \label{fig:vio_adjusted_overlays}
\end{figure}

Divergences for the combined two test sequences are shown in Table \ref{tab:vio_adjustment_results} and visualized in Figure \ref{fig:vio_adjusted_overlays}. The ``\% Dec." column in Table \ref{tab:vio_adjustment_results} displays the total decrease in divergence from the unadjusted covariances as a percentage of the reduction achieved using ergodic ground-truth. Numbers in the ``Test Set Sampling" column are means and standard deviations of divergences computed from 50 groups of 200 points each from the test sequences. In the interest of space, the last three columns contain percentages for the position and orientations only. The trends are the same as those shown for the 2D localization experiment despite using ergodicity rather than Monte-Carlo simulations to compute ground-truth covariances: the calibration of the ground-truth covariance is within one standard deviation of the calibration of both neural networks and the calibrations of Hypotheses 1 and 2 are inadequate. Once again, when there was very little overlap between a histogram and the $\chi^2$ density, we did not plot the $\chi^2$ density.

\section{Conclusions and Future Work}

We have shown that there exists a learnable map between the uncalibrated estimates of a typical Extended Kalman Filter for VIO and the true, calibrated values. Another conclusion is that the ergodicity assumption is a reasonable way to compute a ``ground-truth" value for covariance for suitable ground-truth motion when Monte-Carlo trials are not possible. Our next step will be to investigate how generalizable these results are. The datasets in this work consisted of several sequences all collected using the same sensor in the same environment. We suspect that the calibration described in this work can be performed once for each sensor and VIO implementation, but generalize across multiple environments. A second direction of research would include using information from a neural network to modify an EKF in the loop in an end-to-end fashion instead of simply adjusting the covariances post-hoc. In other words, the final line of equation \eqref{eq:ekf} would take the form:
\begin{equation}
    \hat P_{k+1} = \phi_w(P_k, \hat x_k)
\end{equation}
where $\phi_w$ is a recurrent network. This raises questions not only of prediction accuracy, but of stability, since the online optimization would create closed-loop dependencies.

\addtolength{\textheight}{-14cm}   %

\bibliographystyle{ieeetran}
\bibliography{references}

\end{document}